\documentclass[runningheads]{llncs}
\usepackage[T1]{fontenc}
\usepackage{graphicx}
\usepackage{booktabs}
\usepackage[misc]{ifsym}
\usepackage{amsmath}
\usepackage{amssymb}
\usepackage{xcolor}
\usepackage{hyperref}
\usepackage{subcaption}
\usepackage{multirow}
\usepackage{tabularx}
\hypersetup{colorlinks=true,linkcolor=black,citecolor=black,urlcolor=blue}

\newcommand{\ghlogo}{\raisebox{-0.2ex}{\includegraphics[height=1.0em]{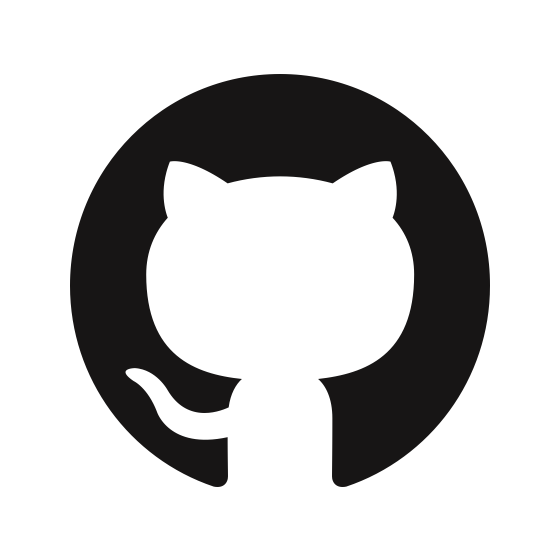}}}
\newcommand{\hflogo}{\raisebox{-0.25ex}{\includegraphics[height=1.0em]{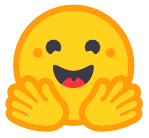}}}

\begin{document}

\title{Demographically-Conditioned Synthetic Medical Images for Bias Mitigation and Bias Detection in Disease Classifiers}

\titlerunning{Demographically-Conditioned Synthetic Chest CT}

\author{Mahmoud Ibrahim\inst{1,2,3}\thanks{Corresponding author: \email{mahmoud.ibrahim@vito.be}} \and
Bart Elen\inst{3} \and
Chang Sun\inst{1,2} \and
G\"okhan Ertaylan\inst{3} \and
Michel Dumontier\inst{1,2}}
\authorrunning{M. Ibrahim et al.}
\institute{Institute of Data Science, Faculty of Science and Engineering, Maastricht University, Maastricht, The Netherlands \and
Department of Advanced Computing Sciences, Faculty of Science and Engineering, Maastricht University, Maastricht, The Netherlands \and
VITO, Belgium\\
\email{mahmoud.ibrahim@vito.be}}

\maketitle

\begin{abstract}
Per-subgroup fairness audits of medical image classifiers face a
sample-size problem: minority subgroups in held-out test sets have so
few samples that the resulting confidence intervals on per-subgroup
performance are wider than the bias the audit is meant to detect. We
argue that a demographically-conditioned synthetic generator can do
both: \emph{mitigate} bias on the training side and \emph{detect} bias
on the evaluation side. Working on COVID-19 chest CT classification
with an end-to-end fine-tuned Stable Diffusion~2.1 generator, we make
two findings. \textbf{For bias mitigation (training)}, a
demographically-balanced synthetic cohort is most useful as a
pretraining prior, not as joint augmentation: with the same fixed
data, sequential pretraining followed by fine-tuning substantially
outperforms joint augmentation, and the resulting classifier surpasses
the full-real baseline at $\sim$$100\times$ real-data efficiency.
\textbf{For bias detection (evaluation)}, across five synthetic
minority cohorts and five classifier seeds, the synthetic estimator
reproduces the subgroup ranking of a well-powered real oracle
(Spearman $\rho = 1.00$ on MCC and Recall) and gives the more reliable
per-cell estimate where the small real test set runs out of samples.
The synthetic cohort is therefore most useful in exactly the cells that
fairness audits care about, as both a fix for and a measure of subgroup
bias.

\end{abstract}

\begin{center}
\small
\ghlogo~\href{https://github.com/mahmoudibrahim98/synthetic-fairness}{\texttt{Code}}
\quad
\hflogo~\href{https://huggingface.co/mahmoudibra98/covid-ct-sd21}{\texttt{Model weights}}
\end{center}

\section{Introduction}
\label{sec:intro}

Per-subgroup fairness audits of medical image classifiers face a
sample-size problem. Held-out real test sets are dominated by majority
demographics, leaving minority subgroups with only a few dozen slices
per (sex, age) cell. At that sample size, per-subgroup confidence
intervals are wider than the bias the audit is meant to detect
\cite{seyyed2020chexclusion,seyyed2021underdiagnosis,yang2023limits},
and an observed disparity cannot be distinguished from sampling noise.
The same imbalance drives the bias being audited: training data also
under-represents those cells, pushing the decision boundary away from
minorities and producing real per-subgroup performance gaps that
aggregate accuracy hides. For COVID-19 chest CT classification, where
real-only models routinely reach $> 0.99$ aggregate ROC-AUC, both
problems show up in the same place: the cells with the fewest patients.

The minority-subgroup problem in medical imaging is therefore
two-sided. On one hand, there is real bias to \emph{mitigate}: the
classifier underperforms on minorities because the training data
under-represents them. On the other, there is bias that is hard to
\emph{detect}: the test set under-represents the same cells, so the
audit cannot tell a real gap apart from sampling noise. A
demographically-conditioned generative model is a natural lever on
both sides of this problem. It can fill the per-cell training gaps
that bias the classifier, and it can produce a synthetic minority test
set large enough to give statistically stable subgroup estimates.
Prior work has explored both directions separately, for chest
radiograph synthesis
\cite{chambon2022roentgen,packhauser2023generation,cho2024medisyn}
and for synthetic-test-set evaluation more broadly
\cite{vanbreugel2023synthetic,vanbreugel2023syng4me,koetzier2024generating}.
What is missing is a practical account of \emph{how} synthetic data is
most useful in each role: as joint augmentation alongside real data,
or as a pretraining prior that the real data later refines? And does
the resulting synthetic cohort actually give a more reliable
subgroup-fairness estimate than the small real test set a practitioner
would typically have?

We address these two questions for COVID-19 chest CT classification.
We fine-tune Stable Diffusion~2.1 \cite{rombach2022highres} end-to-end
on the COVIDx-CT-3A dataset \cite{gunraj2022covidnetct} with
natural-language prompts conditioned on sex, age, and COVID-19 status,
and evaluate the generator on both sides of the problem: a training
sweep over augmentation and two-stage pretraining regimes, and a
five-classifier $\times$ five-cohort subpopulation audit against both
a small held-out real test set and a well-powered real minority oracle.

\subsection{Main contributions}
\label{sec:contributions}

\begin{enumerate}
  \item \textbf{A demographically-conditioned chest CT generator.}
    End-to-end fine-tuning of a general-image latent-diffusion model
    on COVIDx-CT-3A produces a usable demographically-conditioned
    generator, beating a CXR-domain-pretrained baseline on FID while
    matching conditioning accuracy. The generator is the
    common substrate of the two contributions below.
  \item \textbf{Bias mitigation: balanced-synthetic data, used as a
    prior rather than as augmentation, turns a tiny real fine-tune
    into a fair classifier.} Using balanced-synthetic data as a
    pretraining prior, then fine-tuning on $1\,\%$ of the real training
    set, produces a classifier that surpasses the full-real baseline
    on both mean and worst-cell MCC at $100\times$ real-data efficiency. A paired comparison at fixed real
    $+$ synthetic data isolates the \emph{schedule} (sequential
    pretraining followed by fine-tuning), not the data composition, as the source
    of the bias-mitigation lift: the same data combined jointly
    underperforms the same data applied sequentially by a large margin
    on worst-cell fairness.

  \item \textbf{Bias detection: a synthetic minority cohort detects
    subgroup bias more reliably than the small real test, in exactly
    the cells where bias matters most.} Across five hold-out
    classifiers and five synthetic cohorts, the synthetic estimator
    reproduces the subgroup ranking of a well-powered real oracle
    perfectly on MCC and Recall (Spearman $\rho = 1.00$). In the
    minority cells where the small real test has the fewest samples,
    the synthetic estimator's per-cell bias is several times smaller
    than the real test's. In well-sampled cells the small real test is
    the better calibrator, so the two estimators are complementary
    along a sample-size axis.
\end{enumerate}

\section{Related Work}
\label{sec:related}

\subsubsection{Conditional medical image synthesis.}
Latent diffusion models \cite{rombach2022highres,ho2020denoising} have
become the dominant approach for high-resolution medical image
generation. In chest radiography, RoentGen \cite{chambon2022roentgen}
fine-tunes Stable Diffusion on chest X-rays with radiology-report
conditioning; \cite{packhauser2023generation} demonstrates anonymized chest
radiograph generation for downstream classification; MediSyn
\cite{cho2024medisyn} extends to multi-modality medical generation. For
chest CT specifically, COVID-19 image synthesis has used both GANs
\cite{jiang2020covid} and diffusion models. We adopt latent diffusion with
classifier-free guidance \cite{ho2022classifier} and CLIP-based text
conditioning \cite{radford2021learning}.

\subsubsection{Synthetic data for training and evaluation.}
Train-on-synthetic-test-on-real (TSTR) and augmentation studies show that
synthetic data can substitute for or complement real training data when the
generator faithfully captures the target distribution
\cite{esteban2017real,koetzier2024generating}. For minority subgroups
specifically, generator-side techniques such as minority guidance
\cite{um2023dont,sehwag2022generating} and class-balancing diffusion
\cite{qin2023classbalancing} aim to over-sample low-density regions of
conditioning space. On the evaluation side, recent work
\cite{vanbreugel2023synthetic,vanbreugel2023syng4me} has begun to formalize
synthetic test sets as a model-evaluation tool when real evaluation data
is scarce or privacy-restricted. Our augmentation and
sequential pretraining and fine-tuning recipes extend the training line of work by
quantifying the data-efficiency gain from balanced-synthetic data; on the
evaluation side, we frame audit validity as a combined rank-order and
size-gated calibration question on a patient-disjoint subpopulation
regime that controls for the size asymmetry between real-test and
real-minority partitions.

\subsubsection{Fairness and subgroup evaluation in medical AI.}
Subgroup performance gaps have been documented for chest radiograph
classifiers \cite{seyyed2020chexclusion,seyyed2021underdiagnosis}, glaucoma
detection \cite{lin2023evaluate}, and others
\cite{yang2023limits}. Many of these studies are limited by the same
problem we target: real minority sample sizes are too small for reliable
per-subgroup estimation. Synthetic minorities offer a complement when used
carefully, but as we show only as a rank-order identifier, not as an
absolute estimator.

\section{Methods}
\label{sec:methods}

\subsection{Dataset and patient-disjoint partitions}
\label{sec:dataset}
We use the COVIDx-CT-3A release of the COVID-Net family
\cite{gunraj2022covidnetct}, retaining only the Variant-A subset with
confirmed labels and excluding pneumonia cases to preserve a binary
\textit{Normal} vs \textit{COVID-19} task. After preprocessing this yields
$168{,}694$ axial chest CT slices from $1{,}718$ unique patients, with
patient-level sex (F/M) and age categorized into five groups
(Under 20, 20--40, 40--60, 60--80, Over 80). All splits are
\emph{patient-disjoint} to prevent leakage and stratified on the joint
(sex, age, label) distribution.

\begin{figure}[!htbp]
  \centering
  \includegraphics[width=\textwidth]{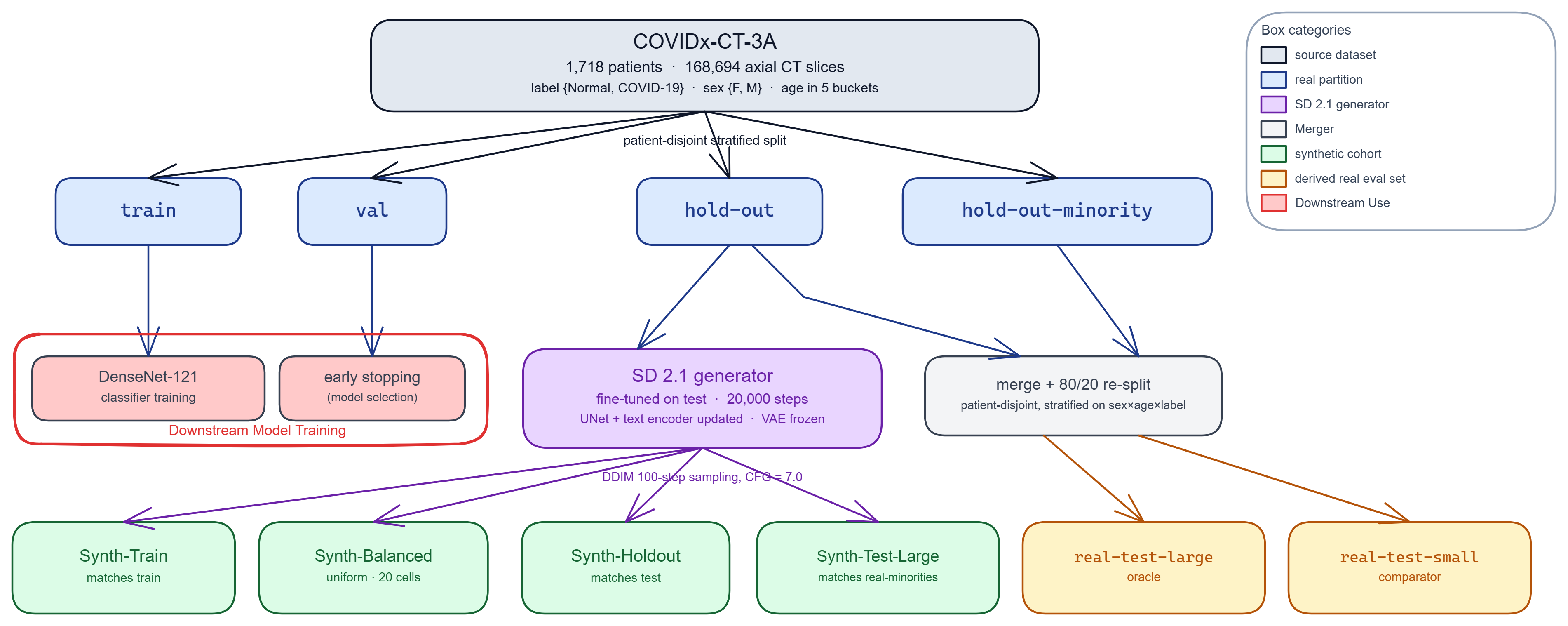}
  \caption{\textbf{Data-provenance map.} The four COVIDx-CT-3A
  patient-disjoint partitions, the two derived per-subgroup evaluation
  sets (\texttt{real-test-large}, \texttt{real-test-small}) obtained by
  merging and re-splitting \texttt{hold-out}\,$\cup$\,\texttt{hold-out-minority},
  the SD~2.1 fine-tuned generator, and the four synthetic cohorts
  sampled from it. Exact counts and the downstream use each set feeds
  are given in Table~\ref{tab:splits}.}
  \label{fig:splits}
\end{figure}

We split the source data into four base partitions
(Figure~\ref{fig:splits}, Table~\ref{tab:splits}): \texttt{train},
\texttt{val}, \texttt{hold-out}, and \texttt{hold-out-minority}.
\texttt{train} and \texttt{val} have their conventional roles.
\texttt{hold-out} ($62{,}829$ slices) matches the joint (sex, age, label)
distribution of \texttt{train}/\texttt{val} and is the held-back real
reference; the name marks its dual role as the generator's training
data (Section~\ref{sec:generator}). \texttt{hold-out-minority}
($24{,}969$ slices) is a separate partition deliberately enriched for
minority (sex, age) cells, set aside so the per-subgroup audit reaches
a per-cell sample size \texttt{hold-out} alone cannot deliver.
Aggregate experiments
(Sections~\ref{sec:results_generator}--\ref{sec:results_transfer}) use
the standard three-way \texttt{train}/\texttt{val}/\texttt{hold-out}
split. The per-subgroup audit (Section~\ref{sec:audit}) pools
\texttt{hold-out} with \texttt{hold-out-minority} and re-splits the
union $80/20$ at the patient level under the same stratification,
yielding the well-powered minority oracle \texttt{real-test-large}
($71{,}037$ slices) and the small comparator \texttt{real-test-small}
($16{,}761$ slices). \texttt{real-test-small} emulates the deployment
scenario: a small held-out audit set with realistic minority sparsity,
the kind of test set a practitioner would actually have.
\texttt{real-test-large} is the ground-truth oracle against which both
\texttt{real-test-small} and the synthetic proxy are validated. The
$\sim$$4\times$ size asymmetry between them is the experimental
variable the audit in Section~\ref{sec:audit} is built around.

\begin{table}[t]
  \centering
  \caption{Real partitions paired with their size-matched synthetic
  mirrors. All real splits are patient-disjoint; synthetic cohorts are
  sampled per-slice from the fine-tuned SD~2.1 generator (DDIM 100-step,
  CFG~7.0). 
  Synth-Balanced is uniform version of real data over the
  $20$ sex$\times$age$\times$label cells, $2{,}000$ slices each).}
  \label{tab:splits}
  \setlength{\tabcolsep}{4pt}
  \small
  \begin{tabularx}{\textwidth}{@{}lrrlX@{}}
    \toprule
    Real set & Slices & Pat. & Synthetic mirror & Role / downstream use \\
    \midrule
    \texttt{train}                 & 58{,}499 & 588 & Synth-Train     & classifier training; TSTR-Matched, FT-Matched \\
    \texttt{val}                   & 22{,}397 & 197 & --              & early stopping \\
    \texttt{hold-out}              & 62{,}829 & 644 & Synth-Holdout   & generator fine-tuning; threshold sweep (Fig.~\ref{fig:thresh_sweep}) \\
    \texttt{hold-out-minority}     & 24{,}969 & 289 & --              & merged with \texttt{hold-out} for the audit re-split \\
    \addlinespace
    \texttt{real-test-large}\,$^{\dagger}$ & 71{,}037 & 746 & Synth-Test-Large & per-subgroup oracle / proxy (\S\ref{sec:audit}) \\
    \texttt{real-test-small}\,$^{\dagger}$ & 16{,}761 & 187 & Synth-Test-Small & Table~\ref{tab:agg_realsynth} aggregate pair \\
    \addlinespace
    --                                     & 40{,}000 & --  & Synth-Balanced   & AUG; FT-Balanced (uniform over $20$ cells) \\
    \bottomrule
  \end{tabularx}
  \\[2pt]
  \footnotesize
  $^{\dagger}$\,Derived by patient-disjoint $80/20$ re-split of
  \texttt{hold-out}\,$\cup$\,\texttt{hold-out-minority}
  ($87{,}798$ slices / $933$ patients). Source totals:
  $1{,}718$ patients, $168{,}694$ slices.
\end{table}

\subsection{Generators}
\label{sec:generator}

\paragraph{Baseline (CXR-pretrained).}
Following prior work on chest radiograph synthesis with latent
diffusion, we adopt as a baseline the publicly available MONAI
implementation \cite{monai_cxr_ldm,warvito_cxr_ldm} of a latent
diffusion model pre-trained on MIMIC-CXR
\cite{johnson2019mimiccxr}. The model has the standard three
components (U-Net denoiser, VAE, and text encoder), and we fine-tune
the U-Net and VAE on COVIDx-CT-3A while keeping the text encoder
frozen, matching the recipe of the public release.

\paragraph{Ours (SD~2.1 fine-tuned end-to-end).}
Our generator instead starts from publicly released Stable
Diffusion~2.1 \cite{rombach2022highres} weights and fine-tunes the
U-Net denoiser and the CLIP text encoder \cite{radford2021learning}
end-to-end on the \texttt{hold-out} partition for $20{,}000$ optimization
steps; only the variational autoencoder is held fixed. The VAE
compresses $512 \times 512$ inputs to $64 \times 64 \times 4$ latents.
The two design choices are ablated in Table~\ref{tab:fid}.

Demographic and diagnostic attributes are encoded as a natural-language
prompt of the form
  \texttt{"\{age\} year old \{SEX\}.~\{COVID-19 findings on / Normal\} chest CT scan"}
which is tokenized and embedded by the CLIP text encoder; the resulting
77-token embeddings condition the U-Net via cross-attention.
Classifier-free guidance is implemented with $10\,\%$ null-prompt dropout
during training \cite{ho2022classifier}. Inference uses the DDIM scheduler
with $100$ steps and a guidance scale of $7.0$. Training is performed in
mixed bfloat16 precision with gradient checkpointing across six
$40$\,GB A100 GPUs.

\subsection{Synthetic cohorts}
\label{sec:cohorts}
From the fine-tuned generator we sample the five synthetic cohorts in
Table~\ref{tab:splits}: \textbf{Synth-Train}, \textbf{Synth-Holdout}, and \textbf{Synth-Test-Small}
mirror the joint demographic distribution of their real counterparts;
\textbf{Synth-Test-Large} mirrors the well-powered \texttt{real-test-large}
oracle and is the audit proxy in Section~\ref{sec:audit};
\textbf{Synth-Balanced} samples uniformly across the $20$
sex$\times$age$\times$label cells ($2{,}000$ slices each), providing
the demographically-flat pool used by the AUG and FT-Balanced regimes
(Section~\ref{sec:classifier}, Table~\ref{tab:regimes}).

\subsection{Downstream classifier regimes}
\label{sec:classifier}
All downstream classifiers share the same architecture and optimization
recipe: a DenseNet-121 binary classifier (\textit{Normal} vs \textit{COVID-19})
initialized from ImageNet weights, trained with Focal Loss
($\gamma = 2.0$), Adam (learning rate $1\!\times\! 10^{-4}$, batch
size $64$), and \textit{ReduceLROnPlateau} on $1 - $val-F1. Early stopping
is performed by keeping the epoch with the highest macro F1 on the real
\texttt{val} partition.

We train $22$ regimes, listed individually in Table~\ref{tab:regimes}
and grouped into four families: (i) real-only baselines that establish
the upper bound a real-data classifier reaches in the
demographically-biased and the demographically-balanced settings;
(ii) synthetic-only baselines that establish the TSTR lower bound;
(iii) a six-point augmentation sweep that mixes a fixed
$40{,}000$-slice balanced-synthetic pool with growing fractions of
real \texttt{train}; and (iv) two six-point two-stage sweeps that
pretrain on a synthetic cohort and then fine-tune on a fraction of
real \texttt{train}, contrasting balanced versus distribution-matched
pretraining. Fine-tuning uses learning rate $1\!\times\! 10^{-5}$ (one
order of magnitude lower than pretraining) for $15$ epochs.

\begin{table}[t]
\caption{The $22$ downstream classifier regimes, one row per family.
For two-stage families the data spec reads
\emph{pretrain $\to$ fine-tune}; for single-stage families it is the
direct training data. Sweep families enumerate the six sweep points
inline.
Acronyms: TRTR = train-on-real-test-on-real,
TSTR = train-on-synthetic-test-on-real, AUG = augmentation,
FT = fine-tuning.}
\label{tab:regimes}
\centering
\small
\setlength{\tabcolsep}{4pt}
\begin{tabularx}{\textwidth}{@{}llX@{}}
\toprule
Family ($n$) & Regime & Data spec \\
\midrule
Real-only ($2$)         & TRTR-Biased        & real \texttt{train} (full $58{,}499$) \\
                        & TRTR-Balanced      & real \texttt{train}, class-balanced subset \\
Synthetic-only ($2$)    & TSTR-Matched       & Synth-Train ($58{,}499$, matches \texttt{train}) \\
                        & TSTR-Balanced      & Synth-Balanced ($40{,}000$, uniform) \\
Augmentation ($6$)      & AUG~$@$~N          & real$_N$ $\cup$ Synth-Balanced ($40$k) \\
Balanced pretrain ($6$) & FT-Balanced~$@$~p\%& Synth-Balanced $\to$ real$_{p\%}$ \\
Matched pretrain ($6$)  & FT-Matched~$@$~p\% & Synth-Train $\to$ real$_{p\%}$ \\
\bottomrule
\end{tabularx}
\\[2pt]
\footnotesize
Sweep grids: $N \in \{500, 2\text{k}, 10\text{k}, 20\text{k}, 40\text{k}, \text{full}\}$ for AUG;
$p \in \{1, 10, 30, 50, 70, 100\}$ for both FT families.
\end{table}

\subsection{Evaluation metrics and audit setup}
\label{sec:metrics}
\paragraph{Per-regime metrics.}
Our primary metric across the regime sweep is Matthews Correlation
Coefficient (MCC), robust to per-cell class imbalance and reported as
mean and worst-cell across the $10$ sex$\times$age subgroups. We
additionally report ROC-AUC, Recall, Accuracy, macro-F1, and Precision.
For the subpopulation evaluation we use the same $10$ subgroups,
pooling Normal and COVID labels within each, and report mean and
$95\,\%$ CI over the $5$ classifier seeds ($\times\,5$ synthetic
cohorts for synthetic estimators).

\paragraph{Subpopulation audit.}
The audit (\S\ref{sec:audit}) studies how to reliably evaluate  per-subgroup performance. To do so, we fix a real-data-trained classifier (TRTR-Biased) and
evaluate it on three test cohorts over the same patient-disjoint
subgroups: \texttt{real-test-large} (the well-powered \emph{oracle}),
\texttt{real-test-small} (a small real \emph{comparator}, the kind of
test set a practitioner would actually have at deployment), and
\texttt{synth-test-large} (a \emph{synthetic proxy} sampled from the
conditional generator at the oracle's size and demographic
distribution). The synthetic proxy is the candidate solution we
propose for the small-real-test-set problem: when real minority
slices are scarce, we sample a large, demographically-balanced
synthetic cohort to take their place. The audit then asks how closely
the proxy reproduces the oracle's per-subgroup pattern along two axes.
\emph{Magnitude agreement} is the per-subgroup mean F1 and MCC, with
absolute bias $|m_{\text{proxy}} - m_{\text{oracle}}|$ on MCC and F1.
\emph{Rank agreement} is Spearman $\rho$ between the proxy's
per-subgroup ranking and the oracle's on MCC, Recall, and AUC-PR, with
a $95\,\%$ bootstrap CI.

\section{Experiments and Results}
\label{sec:results}

\subsection{Generator quality}
\label{sec:results_generator}
Aggregate FID against the matching real partitions is reported in
Table~\ref{tab:fid}, alongside cohort-level conditioning accuracy and
COVID-19 classifier AUC. The fine-tuned SD~2.1 generator achieves
FID-Inception $50.14$ and FID-RadImageNet $4.68$ on the
\texttt{hold-out} partition, improving on a CXR-domain-pretrained baseline by
$35\,\%$ on FID-Inception and $32\,\%$ on FID-RadImageNet. Demographic conditioning accuracy on the
synthetic cohorts (measured by separately trained sex and age-group
classifiers) remains $\geq 0.84$ for sex and $\geq 0.33$ for the
five-class age problem (chance $0.20$); a separately trained COVID
classifier achieves AUC $> 0.99$ on the synthetic cohorts, confirming
that diagnosis conditioning is preserved.

\begin{table}[t]
\caption{Aggregate generator fidelity. Our SD~2.1 (fine-tuned
end-to-end) versus the CXR-pretrained latent-diffusion baseline.
\emph{Sex/Age acc.}: conditioning accuracy of separately-trained sex
and age-group classifiers on the synthetic cohorts. \emph{COVID AUC}:
ROC-AUC of a separately-trained COVID classifier on the synthetic
cohorts.}
\label{tab:fid}
\centering
\small
\setlength{\tabcolsep}{4pt}
\begin{tabularx}{\textwidth}{@{}Xcccccc@{}}
\toprule
Cohort & $n$ & FID-Inc.\,$\downarrow$ & FID-Rad.\,$\downarrow$
       & Sex\,$\uparrow$ & Age\,$\uparrow$ & COVID AUC\,$\uparrow$ \\
\midrule
CXR baseline             & 62.8k & 77.04           & 6.93          & --    & --    & --    \\
\midrule
Ours / Synth-Holdout     & 62.8k & \textbf{50.14}  & \textbf{4.68} & 0.854 & 0.332 & 0.995 \\
Ours / Synth-Train       & 58.5k & 50.37           & 4.54          & 0.854 & 0.348 & 0.994 \\
Ours / Synth-Test-Large  & 71.0k & 47.41           & 4.12          & 0.842 & 0.371 & 0.990 \\
\bottomrule
\end{tabularx}
\end{table}

\subsection{Bias mitigation I: Real+synthetic augmentation improves real-data classifiers}
\label{sec:results_augmentation}

\begin{figure}[!htbp]
\centering
\includegraphics[width=\textwidth]{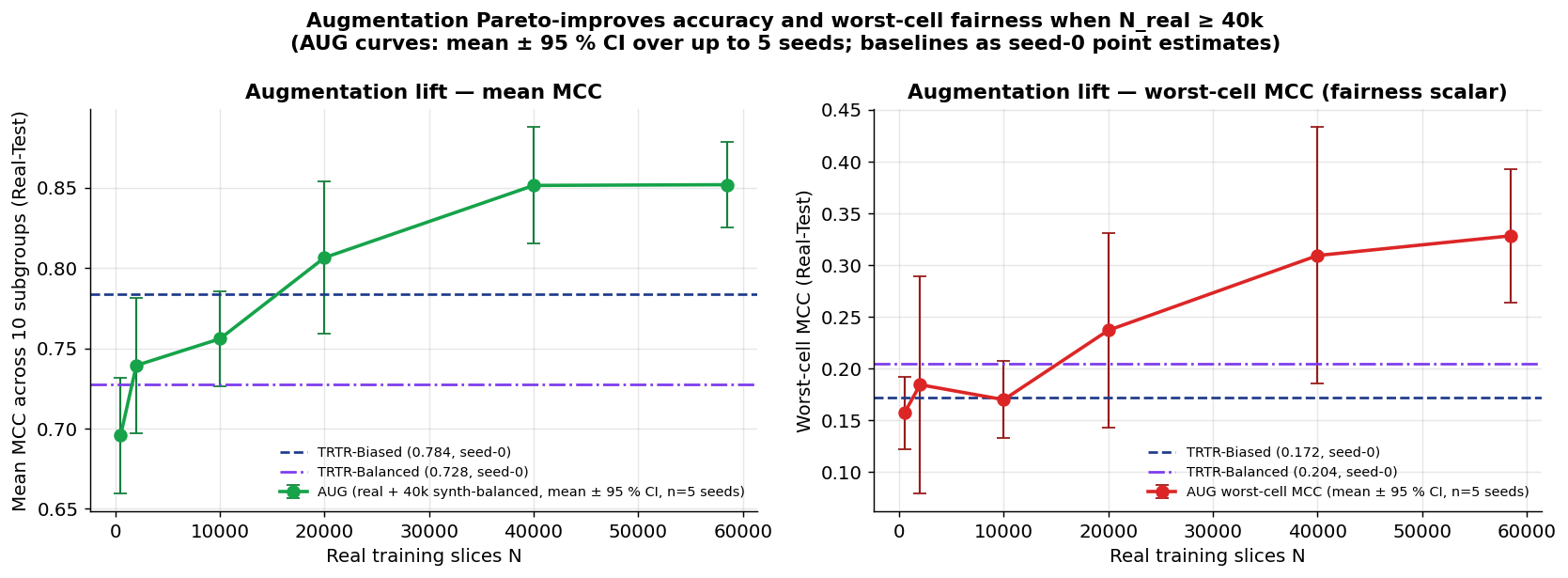}
\caption{\textbf{Augmentation lift on Real-Test (5-seed CIs).} Mean
(left) and worst-cell (right) MCC across $10$ subgroups, as a function
of real training-set size $N$. AUG curves: mean $\pm$ $95\,\%$ CI over
five classifier seeds. Real-only baselines (TRTR-Biased $0.784$ / $0.172$,
TRTR-Balanced $0.728$ / $0.204$) shown as horizontals.
AUG~$@$~$N\!\ge\!40{,}000$ exceeds both baselines on \emph{both} metrics.}
\label{fig:aug}
\end{figure}

We first evaluate whether augmenting real training data with
demographically-balanced synthetic slices improves the resulting
classifier's per-subgroup performance. The AUG sweep adds
$40{,}000$ Synth-Balanced slices to varying real-data quantities
ranging from $500$ to the full $58{,}499$. We compare these to the
two real-only baselines: TRTR-Biased (full real \texttt{train}, mean
MCC $0.784$, worst-cell $0.172$) and TRTR-Balanced (class-balanced
subset of \texttt{train}, mean MCC $0.728$, worst-cell $0.204$).
TRTR-Balanced is the fairness-conscious alternative within real-only
training.
Figure~\ref{fig:aug} shows the mean and worst-cell MCC across the
$10$ sex--age subgroups on the held-out real test set.

Two observations emerge. (i) Augmentation is most useful in the
\emph{real-rich} regime: AUG $@$ N$=40$k and AUG $@$ N$=$full both
surpass \emph{both} real-only baselines on every metric (mean MCC
$0.852$ and $0.852$ respectively, vs $0.78$ TRTR-Biased and $0.73$
TRTR-Balanced; worst-cell MCC $0.31$, $0.33$ vs $0.17$ and $0.20$).
(ii) At very small real $N$ (e.g.~$500$ real slices), adding the same
$40{,}000$ synthetic slices \emph{harms} performance, as the synthetic
distribution shift dominates the small real signal. The crossover
above TRTR-Biased on mean MCC occurs around $N_{\text{real}} =
20{,}000$. 

The augmentation regime that best balances accuracy and worst-cell
fairness is AUG $@$ N$=40$k: it improves worst-cell MCC by $+0.137$
over TRTR-Biased and $+0.105$ over TRTR-Balanced without sacrificing
mean MCC against either. This is a Pareto-improving modification of
the real training pipeline at modest synthetic cost.

\subsection{Bias mitigation II: Two-stage synthetic pretraining: $100\times$ real-data efficiency}
\label{sec:results_transfer}

\begin{figure}[!htbp]
\centering
\includegraphics[width=\textwidth]{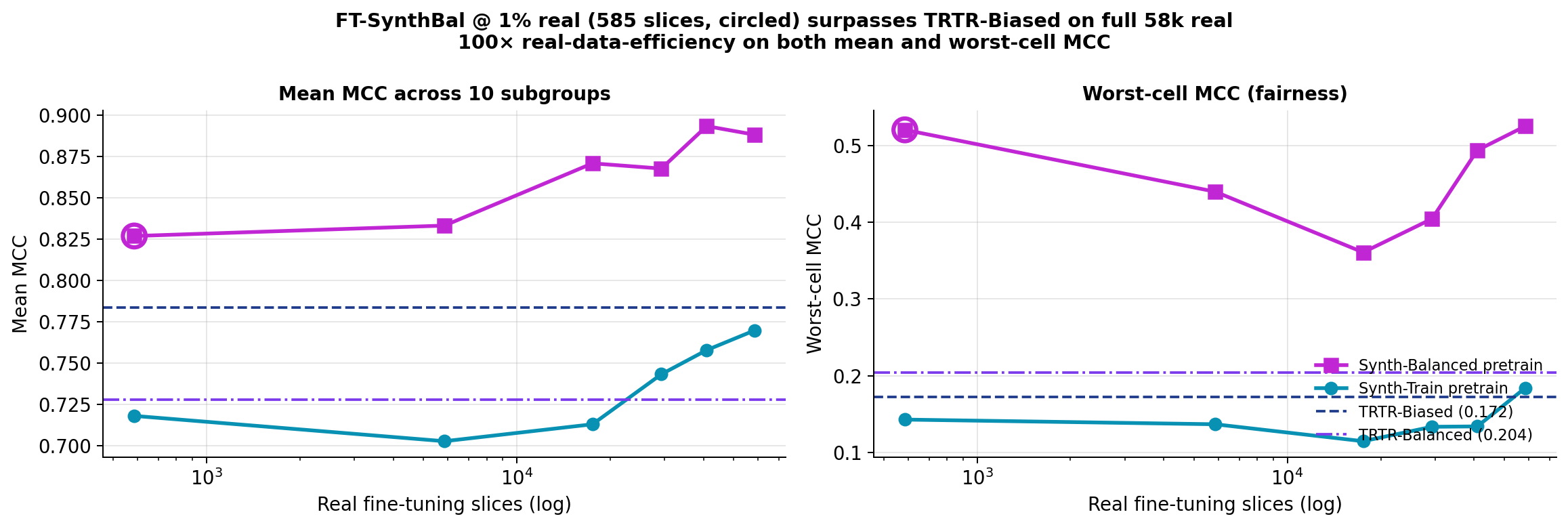}
\caption{\textbf{Two-stage synthetic pretraining + real fine-tuning.}
Mean (left) and worst-cell (right) MCC on Real-Test, as a function of
real fine-tuning data size (log scale). FT-Balanced and FT-Matched
curves: mean $\pm$ $95\,\%$ CI over five seeds. The circled point marks FT-Balanced at $1\,\%$
real ($585$ slices), which clears both baselines on both metrics
($100\times$ real-data efficiency); FT-Matched stays roughly
$0.12$ MCC below FT-Balanced across the sweep.}
\label{fig:transfer}
\end{figure}

We next test whether \emph{synthetic pretraining} followed by
\emph{real fine-tuning} can substitute for large real training sets. The
two-stage FT sweep pretrains DenseNet-121 on one of two synthetic cohorts
(Synth-Balanced or Synth-Train) for $20$ epochs, then fine-tunes on
varying percentages of the real \texttt{train} partition for $15$ epochs
at $1/10$ the pretraining learning rate.
Figure~\ref{fig:transfer} shows the resulting mean and worst-cell MCC on
the real test set.

The headline finding: \textbf{FT-Balanced $@$ $1\,\%$ real
($585$ slices)} reaches mean MCC $0.83$ and worst-cell MCC $0.52$,
exceeding \emph{both} full-real-data baselines on both metrics
(TRTR-Biased $0.78$ / $0.17$; TRTR-Balanced $0.73$ / $0.20$).
Crucially, FT-Balanced beats even the fairness-conscious
TRTR-Balanced baseline by $+0.10$ mean MCC and $+0.32$ worst-cell
MCC. The \textbf{synthetic demographic balancing} in the pretrain provides a much
larger fairness lift than \textbf{rebalancing the real training set itself},
while not sacrificing aggregate accuracy. Scaling the same regime
from $1\,\%$ to $100\,\%$ real adds only $+0.06$ mean MCC,
indicating that the balanced-synthetic pretraining provides most of
the representational quality. The corresponding fairness gain
(worst-cell MCC three times the TRTR-Biased baseline) is the highest
of any of the $22$ classifier regimes evaluated, including those
trained on the full real dataset.

\emph{Why balanced and not matched-distribution pretraining?}
FT-Matched (the pretraining cohort matched to the real-\texttt{train}
distribution, blue circles in Fig.~\ref{fig:transfer}) underperforms
FT-Balanced across the entire real-data range: matched-distribution
pretraining inherits the real distribution's demographic bias, while
balanced pretraining pre-loads the model with uniform per-cell exposure
that small real fine-tuning preserves. Notably,
FT-Matched stays \emph{below} the real-only TRTR-Biased baseline at
every fine-tuning size, while FT-Balanced stays \emph{above} it
throughout. The choice of pretraining distribution is therefore a
persistent design decision, not a low-data-regime artifact.

\emph{Strategy beats data composition.}
To isolate the effect of the training \emph{strategy} from the
contribution of the data itself, we compare two regimes that use
essentially the same data budget but combine it differently:
AUG~$@$~N$=500$ mixes $500$ real slices with $40{,}000$ Synth-Balanced
slices in a single optimization pass, while FT-Balanced~$@$~1\%
pretrains on the same $40{,}000$ synthetic slices then fine-tunes on
$585$ real (an $85$-slice difference in real budget). Across five
paired classifier seeds, AUG reaches mean MCC of
$0.696 \pm 0.036$,  while FT-Balanced reaches
$\mathbf{0.805 \pm 0.023}$ : a delta of $+0.109 \pm 0.031$. In terms of fairness, 
AUG reaches  worst-cell MCC of $0.157 \pm 0.035$, while FT-Balanced reaches
$\mathbf{0.524 \pm 0.021}$: a $3.3\times$ improvement on the fairness scalar with a  delta
of  $+0.367 \pm 0.040$ worst-cell. The training schedule, not the data, is the source of the lift: the synthetic data must be applied as a representation
prior and the small real set as a decision-boundary refinement, not as
two streams of training signal pooled together.

\subsection{Synthetic cohorts as a test-set proxy}
\label{sec:audit}

We now ask whether synthetic data can serve as an evaluation proxy for
real test data. We answer this at two granularities (aggregate and
per-subgroup) and then show that the answer depends systematically on
subgroup size, giving a clean operating recipe.

\subsubsection{Aggregate equivalence.}
We first compare the same TRTR-Biased classifier on the real test
partition and on a size-matched synth-test cohort drawn to match its
demographic distribution, averaging across five hold-out classifier
seeds. Table~\ref{tab:agg_realsynth} shows ROC-AUC is essentially
identical ($\Delta = -0.0011$) and Recall drifts by under $1\,\%$,
while threshold-sensitive metrics (F1, MCC, Precision) drift by
$1.8$--$3.5$\,pp. All deltas are statistically significant.

\begin{figure}[!htbp]
\centering
\includegraphics[width=\textwidth]{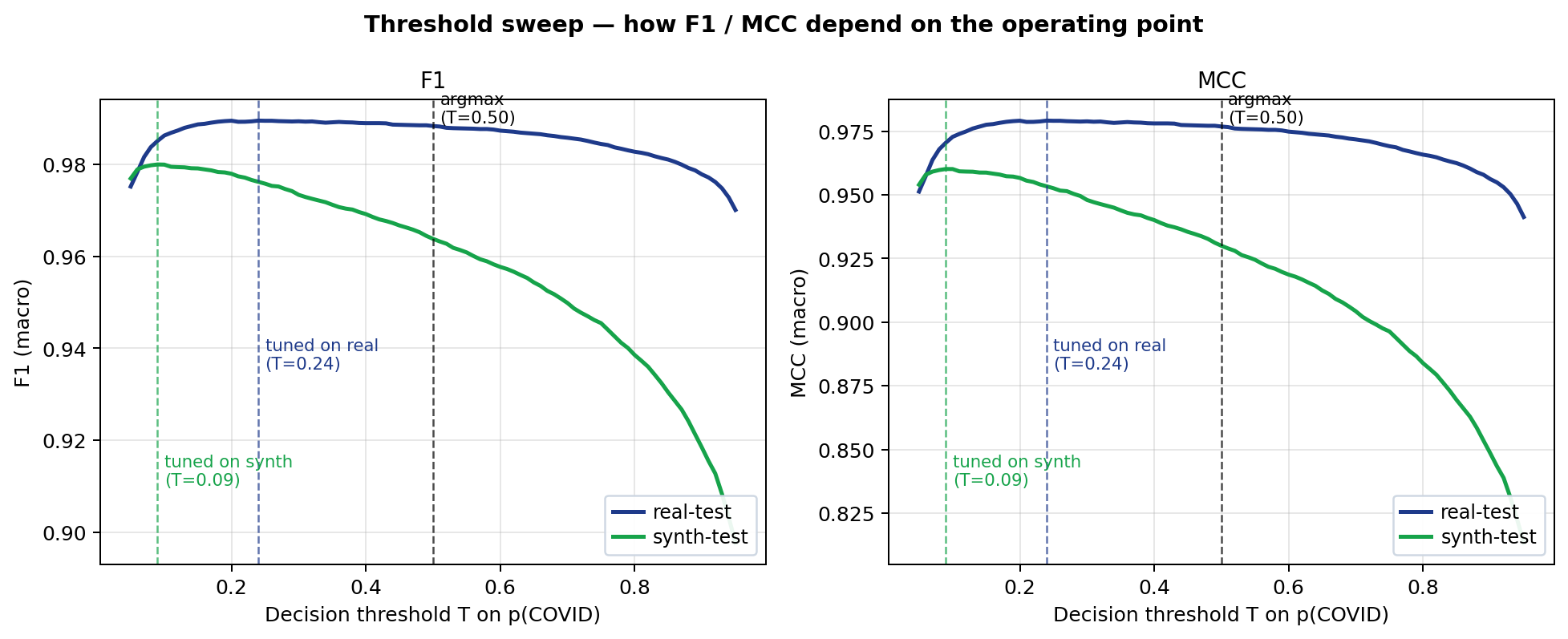}
\caption{\textbf{Threshold sweep on aggregate F1 and MCC.}
Macro-F1 (left) and MCC (right) as a function of the decision
threshold $T$ on $p(\textsc{covid})$, for the TRTR-Biased classifier
on real-test (blue) and synth-test (green). Tuning $T$ on real-val ($T \approx 0.24$)
shrinks the apparent Table~\ref{tab:agg_realsynth} drift by
$\sim$$45\,\%$.}
\label{fig:thresh_sweep}
\end{figure}

\paragraph{Threshold robustness check.}
A threshold sweep over $T \in [0.05, 0.95]$ on
$p(\textsc{covid})$ confirms that the drift in the  threshold-sensitive metrics  is an
operating-point cliff rather than a uniform degradation
(Figure~\ref{fig:thresh_sweep}). At the default $T = 0.5$, the
synth$-$real gap is $-0.025$ F1 and $-0.047$ MCC (synthetic estimate
below the real one). Tuning $T$ on real-val ($T^{*} = 0.24$) and
applying the same threshold to both sets, the deployment-realistic
choice, shrinks this synth$-$real gap to $-0.013$ F1 and $-0.026$ MCC
($\sim$$45\,\%$ reduction). Tuning $T$ on synth-test itself ($T^{*} =
0.09$) shrinks it further to $-0.008$ F1 and $-0.017$ MCC. The
residual $\sim$$1$--$2$\,pp gap is the genuine on-synth degradation;
the rest of the drift in Table~\ref{tab:agg_realsynth} is an artifact
of the default argmax operating point landing on the synth generator's
weakest cell.

\begin{table}[t]
\caption{Aggregate-level comparison with multi-seed $95\,\%$ CIs across
$5$ classifier seeds: Real-Test-Small ($16{,}761$ slices) vs Synth-Test-Small
($16{,}761$ slices) for the real-data-trained TRTR-Biased classifier.
Threshold-free metrics (ROC-AUC, Recall) drift by under $1\,\%$; the
threshold-sensitive metrics (F1, MCC, Precision) drift by
$1.8$--$3.5$\,pp; Accuracy drifts by $0.9$\,pp. All $95\,\%$ CIs
exclude zero.}
\label{tab:agg_realsynth}
\centering
\small
\begin{tabular}{lccc}
\toprule
Metric & Real-Test-Small [$95\,\%$ CI] & Synth-Test-Small [$95\,\%$ CI] & $\Delta$ \\
\midrule
ROC-AUC   & $0.9996 \pm 0.0001$ & $0.9985 \pm 0.0007$ & $-0.0011$ \\
Recall    & $0.9935 \pm 0.0028$ & $0.9847 \pm 0.0089$ & $-0.0088$ \\
Accuracy  & $0.9960 \pm 0.0005$ & $0.9869 \pm 0.0041$ & $-0.0091$ \\
F1        & $0.9919 \pm 0.0009$ & $0.9740 \pm 0.0078$ & $-0.0179$ \\
MCC       & $0.9838 \pm 0.0019$ & $0.9486 \pm 0.0150$ & $-0.0352$ \\
Precision & $0.9903 \pm 0.0020$ & $0.9643 \pm 0.0140$ & $-0.0261$ \\
\bottomrule
\end{tabular}
\end{table}

\subsection{Bias detection: synthetic cohorts as a subgroup auditor.}
Recall the audit setup (\S\ref{sec:metrics}): five real-data trained hold-out classifiers are evaluated on three patient-disjoint test cohorts: 
(i) the well-powered oracle \texttt{real-test-large} ($71{,}037$ slices),
(ii) \texttt{real-test-small} ($16{,}761$, the small real comparator a practitioner
would actually have at deployment), and
(iii)  the synthetic proxy \texttt{synth-test-large} ($71{,}037$ slices matching the well-powered oracle) we propose
as a stand-in when real minority slices are scarce.
The question is
whether the synthetic proxy can reproduce the oracle's per-subgroup
pattern well enough to substitute for it in subgroup auditing.

\begin{figure}[!htbp]
\centering
\includegraphics[width=\textwidth]{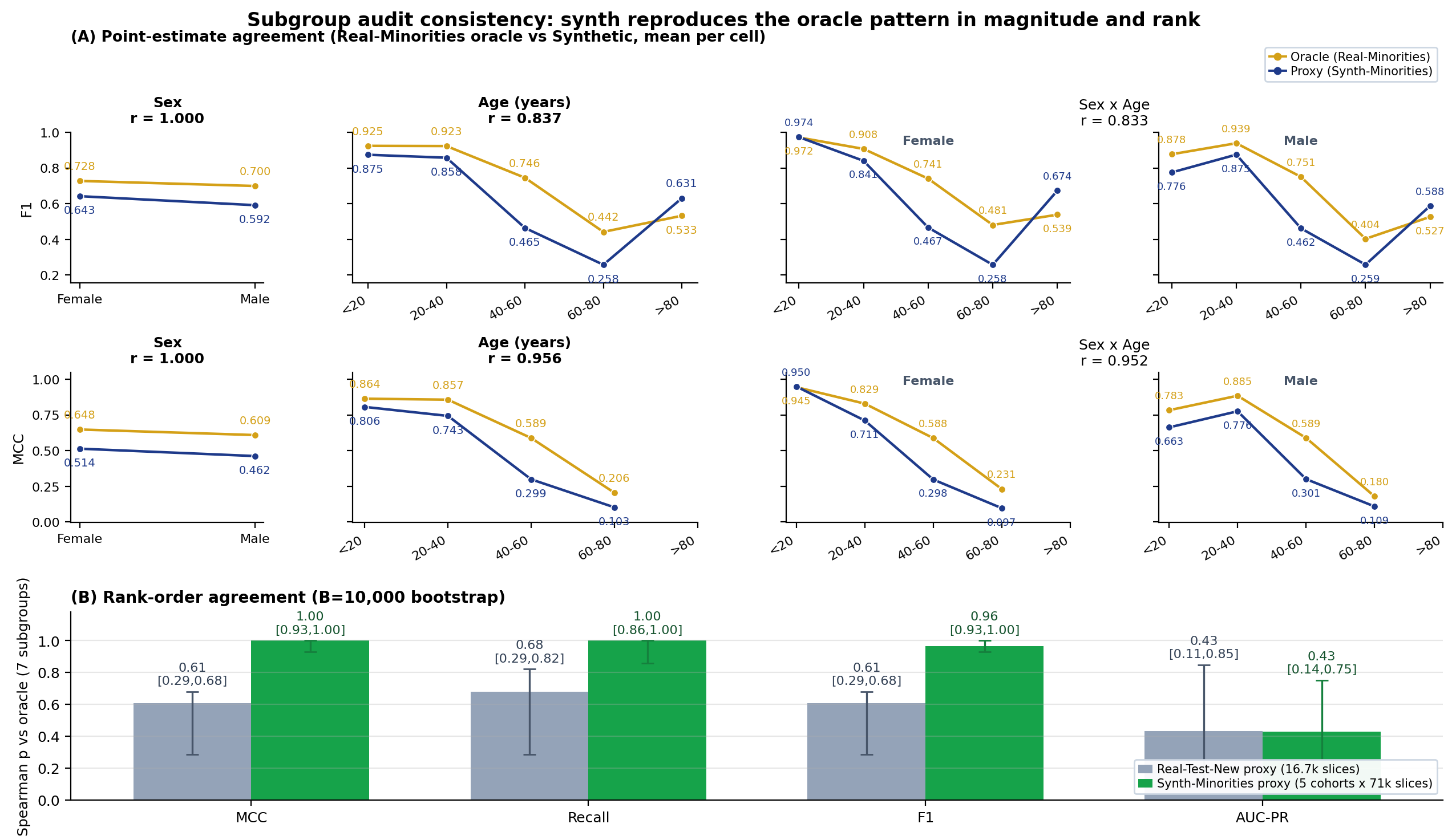}
\caption{\textbf{Subgroup audit consistency.}
\textbf{(A)} Per-subgroup F1 (top) and MCC (bottom) for the
Real-Test-Large oracle (gold) vs Synth-Test-Large proxy (navy, mean
over $5$ cohorts $\times 5$ seeds); per-panel Pearson $r$ over the
cells shown.
\textbf{(B)} Spearman $\rho$ between each proxy's per-subgroup
ranking and the oracle's, $95\,\%$ bootstrap CIs ($7$ evaluable
subgroups, $B=10{,}000$).}
\label{fig:audit_consistency}
\end{figure}

\paragraph{Magnitude agreement (Figure~\ref{fig:audit_consistency}A).}
Plotting mean F1 and MCC for the synthetic estimator and the
\texttt{real-test-large} oracle along each demographic axis, the two lines
track each other on every axis we evaluate: Pearson
$r = 1.00$ across sex, $r = 0.84$ (F1) / $0.96$ (MCC) across the five
age bins, and $r = 0.83$ (F1) / $0.95$ (MCC) across the ten joint
sex--age cells. The $95\,\%$ CIs across the $5$ classifier seeds  are tight for the synthetic estimator across every cell, while
the \texttt{real-test-small} estimator has wide CIs in minority cells
(e.g.~F-Under-20, F-Over-80) where the small real test has fewer than
$200$ slices. The synthetic estimator is therefore both unbiased in
shape and more precise per cell.

\paragraph{Rank agreement (Figure~\ref{fig:audit_consistency}B).}
The Spearman counterpart, computed between each proxy estimator and
the oracle's per-subgroup mean, shows that \texttt{synth-test-large}
achieves perfect rank agreement with the oracle on both MCC and
Recall: $\rho = 1.00$, with $95\,\%$ bootstrap CIs $[0.93, 1.00]$ and
$[0.86, 1.00]$, respectively. This exceeds \texttt{real-test-small}'s rank
agreement on the same subgroups ($\rho = 0.61$, $[0.29, 0.68]$ for
MCC; $\rho = 0.68$, $[0.29, 0.82]$ for Recall). On AUC-PR, rank
correlation is weak \emph{for both} estimators ($\rho = 0.43$,
$[0.14, 0.75]$ for \texttt{synth-test-large}; $[0.11, 0.85]$ for \texttt{real-test-small}):
per-cell AUC-PR drifts in the precision--recall trade-off under both
proxies, so AUC-PR cannot reliably rank these subgroups regardless of
the estimator.

\subsubsection{Synthetic estimates win where the real test fails.}
\begin{figure}[!htbp]
\centering
\includegraphics[width=\textwidth]{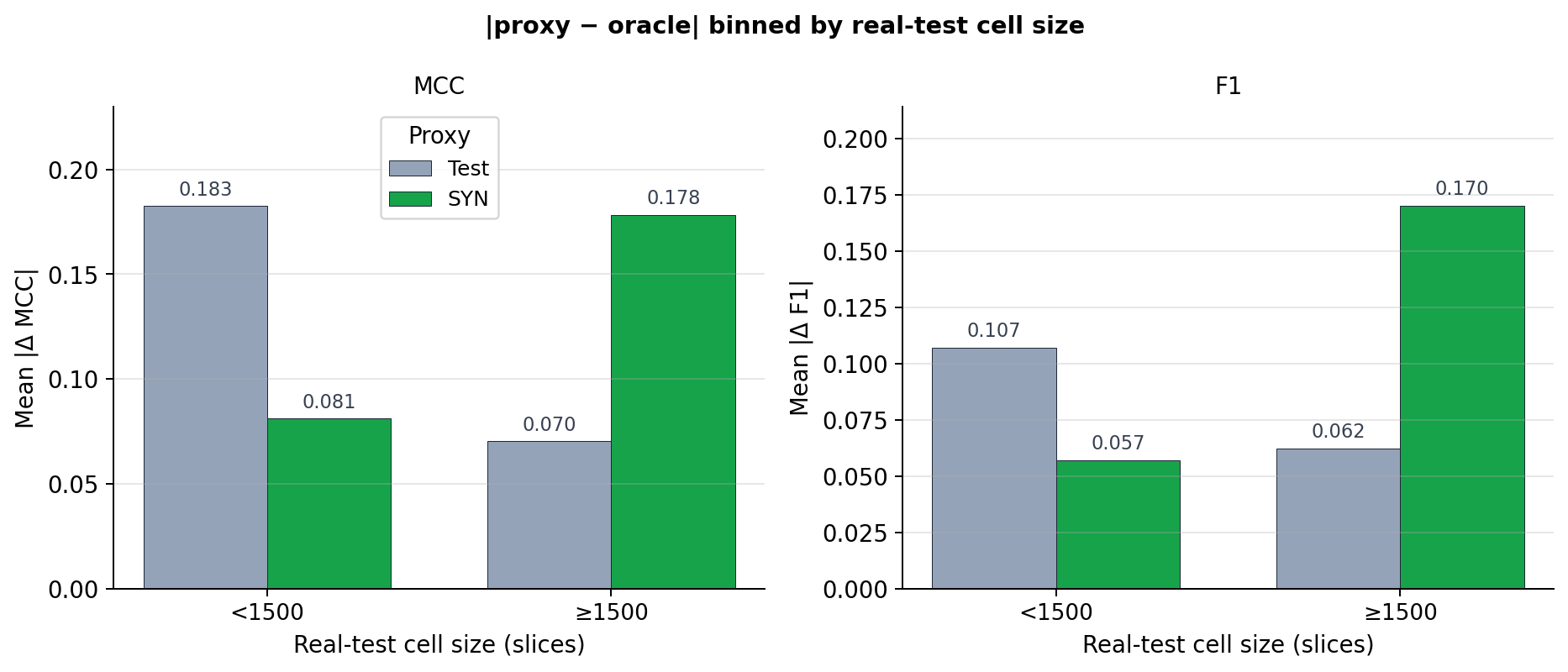}
\caption{\textbf{Calibration error binned by real-test cell size.}
Mean per-cell $|\textnormal{mean}(\textnormal{proxy})
- \textnormal{mean}(\textnormal{oracle})|$ on MCC
(left) and F1 (right), split at $1500$ real-test slices. The order flips between the two
proxies: SYN is $\sim$$2.3\times$ less biased than Test on
under-sampled cells, while Test is $\sim$$2.5\times$ less biased on
adequately-sampled cells. Lower values are better.}
\label{fig:size}
\end{figure}

The synthetic estimator's residual bias is not uniform across
subgroups: it is precisely where the real test runs out of samples
that the synthetic cohort is most useful. Splitting the
subgroups by real-test cell size at $1500$ slices, mean per-cell
$|\textnormal{proxy} - \textnormal{oracle}|$ on MCC flips direction
between the two proxies (Figure~\ref{fig:size}): in the under-sampled
bin ($<1500$ real-test slices) the synthetic cohort's MCC bias is
$0.081$ versus $0.183$ for the real test (a $\sim$$2.3\times$
reduction in calibration error), while in the adequately-sampled bin
($\geq 1500$) the real test is $\sim$$2.5\times$ less biased ($0.070$
vs $0.178$). F1 shows the same crossover ($0.057$ vs $0.107$ under-sampled,
$0.062$ vs $0.170$ adequately-sampled). Per-cell, the gap is even
sharper on the smallest minority: on F-Under-20 the synthetic
estimator's F1 bias is $57\times$ smaller than the real test's
($0.002$ vs $0.114$).
The complementarity is sharpest exactly where each proxy has a
structural advantage: synthetic estimates draw from a large cohort,
so they shine in cells the real test under-samples; the real test,
despite being small overall, has zero distribution shift, so it
dominates whenever the cell itself contains enough real slices for
the estimate to converge.

The two findings (perfect rank agreement; size-dependent calibration)
combine cleanly: a practitioner can use the synthetic estimator to
identify which subgroups warrant attention, then read per-cell
magnitudes from whichever proxy has the lower bias at that cell's real
sample size. The crossover at $\sim$$1500$ real-test slices is a
useful first-cut threshold but is generator-specific (see
\S\ref{sec:discussion}) and should be re-calibrated against a
held-out real reference for each generator and metric.

\section{Discussion}
\label{sec:discussion}

\subsubsection{Why mitigation and detection share a root.}
The two contributions of this paper are not coincidentally compatible.
Both rest on the same property of demographically-conditioned
synthetic data: the generator samples cells that the real data
distribution under-represents. On the training side, that property
fills the per-cell gaps that drive subgroup performance disparities,
so a balanced-synthetic pretrain can stand in for the under-represented
minorities in the training distribution. On the evaluation side, the
same property gives the synthetic minority cohort more samples per
cell than the small real test in the under-represented cells, so the
synthetic estimator inherits a precision advantage where it matters
most. Mitigation and detection are therefore not two
independent uses of the generator; they are two faces of one
structural fix.

\subsubsection{What the $100\times$ data-efficiency claim actually buys.}
Balanced-synthetic pretraining followed by fine-tuning on $585$ real
slices ($1\,\%$ of the real training set) reaches mean MCC $0.83$ and
worst-cell MCC $0.52$, surpassing real-only training on the full
$58{,}499$ slices ($0.78$ / $0.17$) on both metrics. In medical
imaging, labeled data is scarce overall, especially for
minority subgroups. We show how a
balanced resource (synthetic slices) can substitute for the expensive
and rare one (labeled minority data) at a $100\,$:$\,1$ ratio without
losing accuracy or fairness. A single institution with
a few hundred labeled minority slices can therefore train and audit
a fair classifier without requiring the large multi-site cohort.

\subsubsection{Why the bias-mitigation lift lives in the schedule.}
The same logic explains why sequential pretraining and fine-tuning beats
joint augmentation by $+0.367 \pm 0.040$ worst-cell MCC at fixed data
(\S\ref{sec:results_transfer}). In the joint regime, the $500$ real and
$40{,}000$ synthetic slices contribute simultaneously to every
gradient step; the optimizer sees one pooled distribution dominated
$80{:}1$ by synthetic, and the small real signal is unable to correct
generator-specific per-cell drift in the decision boundary. In the
sequential regime, the synthetic data sets up the representation
first, and then the small real fine-tune adjusts only the decision
boundary at $1/10$ the learning rate. The synthetic cohort and the
real cohort therefore play different roles in the pipeline (one builds
the representation, the other refines the boundary), and conflating
them in one training pass discards the structure that makes the prior
useful.

\subsubsection{Threshold protocol for synthetic evaluation.}
The aggregate cliff in Table~\ref{tab:agg_realsynth} forces a choice
of operating point on synthetic test data. Our recommendation: fix $T$
on real-val (or use the deployment default) and report the synth drift
at that $T$. Do not re-tune $T$ on synth: that would overfit the
comparison to the synthetic distribution and disguise the realistic
cost of substitution ($1.3\,\%$ F1 / $2.6\,\%$ MCC for our classifier
at $T = 0.24$). For per-cell audits, the threshold must match across
the real and synth estimators, or the comparison loses meaning. For
purely rank-based use cases, threshold-free metrics sidestep the
question; the real and synth aggregates agree to $\Delta\mathrm{AUC} = 0.0011$.

\subsubsection{Limitations and future work.}
The analyses reported here are based on a single fine-tuned generator
and on a single dataset (COVIDx-CT-3A). Cross-classifier audit, i.e.,
whether synthetic minorities can rank \emph{different classifiers}
consistently, is an open question that we leave for future work.
Privacy-side analyses (nearest-neighbour in feature space, membership
inference) would strengthen claims about synthetic-data deployment.

\section{Conclusion}
\label{sec:conclusion}

Demographically-conditioned synthetic chest CT can both \emph{mitigate}
subgroup bias (in training) and \emph{detect} it (in evaluation).
\emph{On the mitigation side}, a balanced-synthetic cohort serves best
as a representation prior: applied sequentially before a small real
fine-tune, it produces a classifier from $1\,\%$ of the real training
data that surpasses the full-real baseline on both mean and worst-cell
performance. A paired comparison at fixed data isolates the training
schedule, not the data composition, as the source of the lift.
\emph{On the detection side}, the synthetic minority cohort is a
reliable proxy for evaluation and for detecting subgroup bias,
especially in minority cells where labeled real data is scarce. It
matches the well-powered real oracle on both axes the audit cares
about: in magnitude (per-cell F1 and MCC bias several times smaller
than the small real test in under-sampled cells) and in rank
correlation (Spearman $\rho = 1.00$ on MCC and Recall, against
$\rho \approx 0.6$ for the small real test). The two estimators are
complementary along a sample-size axis: the synthetic proxy wins in
under-sampled minority cells, while the small real test wins in
well-sampled ones. Synthetic chest CT is most useful where fairness
audits are weakest, simultaneously as a fix for and a measure of
subgroup bias.

\subsubsection{Code and data availability.}
The training and evaluation code is available at
\ghlogo~\href{https://github.com/mahmoudibrahim98/synthetic-fairness}{\texttt{github.com/mahmoudibrahim98/synthetic-fairness}},
and the fine-tuned generator weights are released at
\hflogo~\href{https://huggingface.co/mahmoudibra98/covid-ct-sd21}{\texttt{huggingface.co/mahmoudibra98/covid-ct-sd21}}.

\begin{credits}
\subsubsection{\ackname} Acknowledgments scrubbed for double-blind review.

\subsubsection{\discintname}
The authors have no competing interests to declare that are
relevant to the content of this article.
\end{credits}


%
%


\begin{thebibliography}{42}

\bibitem{seyyed2020chexclusion}
Seyyed-Kalantari, L., Liu, G., McDermott, M., Chen, I.Y., Ghassemi, M.:
CheXclusion: Fairness gaps in deep chest X-ray classifiers. In: Pacific
Symposium on Biocomputing 2021. pp.~232--243 (2020)

\bibitem{seyyed2021underdiagnosis}
Seyyed-Kalantari, L., Zhang, H., McDermott, M.B.A., Chen, I.Y., Ghassemi, M.:
Underdiagnosis bias of artificial intelligence algorithms applied to chest
radiographs in under-served patient populations. Nature Medicine
\textbf{27}(12), 2176--2182 (2021)

\bibitem{yang2023limits}
Yang, Y., Zhang, H., Gichoya, J.W., Katabi, D., Ghassemi, M.: The limits of
fair medical imaging AI in the wild. arXiv preprint arXiv:2312.10083 (2023)

\bibitem{lin2023evaluate}
Lin, M., Xiao, Y., Hou, B., et al.: Evaluate underdiagnosis and overdiagnosis
bias of deep learning model on primary open-angle glaucoma diagnosis in
under-served populations. In: AMIA Summits on Translational Science
Proceedings. p.~370 (2023)

\bibitem{rombach2022highres}
Rombach, R., Blattmann, A., Lorenz, D., Esser, P., Ommer, B.: High-resolution
image synthesis with latent diffusion models. In: CVPR. pp.~10684--10695 (2022)

\bibitem{ho2020denoising}
Ho, J., Jain, A., Abbeel, P.: Denoising diffusion probabilistic models. In:
NeurIPS. vol.~33, pp.~6840--6851 (2020)

\bibitem{ho2022classifier}
Ho, J., Salimans, T.: Classifier-free diffusion guidance. arXiv preprint
arXiv:2207.12598 (2022)

\bibitem{radford2021learning}
Radford, A., Kim, J.W., Hallacy, C., et al.: Learning transferable visual
models from natural language supervision. arXiv preprint arXiv:2103.00020
(2021)

\bibitem{chambon2022roentgen}
Chambon, P., Bluethgen, C., Delbrouck, J.B., et al.: RoentGen: Vision-language
foundation model for chest X-ray generation. arXiv preprint arXiv:2211.12737
(2022)

\bibitem{packhauser2023generation}
Packh\"{a}user, K., Folle, L., Thamm, F., Maier, A.: Generation of anonymous
chest radiographs using latent diffusion models for training thoracic
abnormality classification systems. In: ISBI. pp.~1--5 (2023)

\bibitem{cho2024medisyn}
Cho, J., Mathur, R., Zakka, C., et al.: MediSyn: A generalist text-guided
latent diffusion model for diverse medical image synthesis. arXiv preprint
arXiv:2405.09806 (2024)

\bibitem{jiang2020covid}
Jiang, Y., Chen, H., Loew, M., Ko, H.: COVID-19 CT image synthesis with a
conditional generative adversarial network. IEEE Journal of Biomedical and
Health Informatics \textbf{25}(2), 441--452 (2020)

\bibitem{esteban2017real}
Esteban, C., Hyland, S.L., R\"{a}tsch, G.: Real-valued (medical) time series
generation with recurrent conditional GANs. arXiv preprint arXiv:1706.02633
(2017)

\bibitem{koetzier2024generating}
Koetzier, L.R., Wu, J., Mastrodicasa, D., et al.: Generating synthetic data
for medical imaging. Radiology \textbf{312}(3), e232471 (2024)

\bibitem{vanbreugel2023synthetic}
van Breugel, B., Seedat, N., Imrie, F., van der Schaar, M.: Can you rely on
your model evaluation? Improving model evaluation with synthetic test data.
In: NeurIPS. vol.~36, pp.~1889--1904 (2023)

\bibitem{vanbreugel2023syng4me}
van Breugel, B., Seedat, N., Imrie, F., van der Schaar, M.: SYNG4ME: Model
evaluation using synthetic test data. OpenReview preprint (2023)

\bibitem{um2023dont}
Um, S., Lee, S., Ye, J.C.: Don't play favorites: Minority guidance for
diffusion models. arXiv preprint arXiv:2301.12334 (2023)

\bibitem{sehwag2022generating}
Sehwag, V., Hazirbas, C., Gordo, A., Ozgenel, F., Canton, C.: Generating
high fidelity data from low-density regions using diffusion models. In: CVPR.
pp.~11492--11501 (2022)

\bibitem{qin2023classbalancing}
Qin, Y., Zheng, H., Yao, J., Zhou, M., Zhang, Y.: Class-balancing diffusion
models. In: CVPR. pp.~18434--18443 (2023)

\bibitem{gunraj2022covidnetct}
Gunraj, H., Sabri, A., Koff, D., Wong, A.: COVID-Net CT-2: Enhanced
deep neural networks for detection of COVID-19 from chest CT images
through bigger, more diverse learning. Frontiers in Medicine
\textbf{8}, 729287 (2022)

\bibitem{johnson2019mimiccxr}
Johnson, A.E.W., Pollard, T.J., Berkowitz, S.J., Greenbaum, N.R.,
Lungren, M.P., Deng, C., Mark, R.G., Horng, S.: MIMIC-CXR, a
de-identified publicly available database of chest radiographs with
free-text reports. Scientific Data \textbf{6}(1), 317 (2019)

\bibitem{monai_cxr_ldm}
MONAI Generative Models: Chest X-ray with Latent Diffusion Models.
GitHub repository,
\url{https://github.com/Project-MONAI/GenerativeModels/tree/7428fce/model-zoo/models/cxr_image_synthesis_latent_diffusion_model}
(2023), last accessed 2026-05-30.

\bibitem{warvito_cxr_ldm}
Pinaya, W.H.L. (Warvito): Latent diffusion models for chest X-ray
generation using MONAI Generative Models. GitHub repository,
\url{https://github.com/Warvito/generative_chestxray} (2023), last
accessed 2026-05-30.

\end{thebibliography}
\end{document}